\documentclass[10pt,twocolumn,letterpaper]{article}

\usepackage[pagenumbers]{cvpr} 

\usepackage{graphicx}
\usepackage{amsmath}
\usepackage{amssymb}
\usepackage{booktabs}
\usepackage{tabularx}
\usepackage{multirow}
\usepackage[table]{xcolor}
\usepackage{capt-of}
\usepackage[symbol]{footmisc}
\usepackage[pagebackref,breaklinks,colorlinks]{hyperref}

\usepackage[capitalize]{cleveref}
\crefname{section}{Sec.}{Secs.}
\Crefname{section}{Section}{Sections}
\Crefname{table}{Table}{Tables}
\crefname{table}{Tab.}{Tabs.}

\usepackage[normalem]{ulem} 

\newcommand{\tinysection}[1]{\noindent\textbf{#1}\hspace{3pt}}

\newcommand{\ignore}[1]{}

\definecolor{best}{RGB}{255, 204, 153}
\definecolor{second}{RGB}{255, 255, 153}
\definecolor{third}{RGB}{255, 255, 255}

\def\cvprPaperID{4980} 
\def\confName{CVPR}
\def\confYear{2022}

\begin{document}

\title{Deblur-NeRF: Neural Radiance Fields from Blurry Images}




\author{Li Ma$^{1}$\footnotemark[1] \qquad Xiaoyu Li$^{2}$ \qquad Jing Liao$^{3}$ \qquad Qi Zhang$^2$ \qquad Xuan Wang$^2$ \\ 
Jue Wang$^2$ \qquad Pedro V. Sander$^{1}$ \vspace{5pt}\\
$^{1}$The Hong Kong University of Science and Technology \\ $^{2}$Tencent AI Lab \qquad $^{3}$City University of Hong Kong \\
}


\newlength{\figtosubcap}
\setlength{\figtosubcap}{-12pt}
\newlength{\figtocap}
\setlength{\figtocap}{0pt}
\newlength{\figtocapexp}
\setlength{\figtocapexp}{-5pt}
\newlength{\captotext}
\setlength{\captotext}{-5pt}
\newlength{\figtofig}
\setlength{\figtofig}{-11pt}
\newlength{\figtocapcamp}
\setlength{\figtocapcamp}{-8pt}
\newlength{\captotextcamp}
\setlength{\captotextcamp}{-2pt}
\twocolumn[{%
\renewcommand\twocolumn[1][]{#1}%
\maketitle

\vspace{-1cm}
\begin{center}
    \centering
\captionsetup{type=figure}
  \rotatebox[origin=c]{90}{\makebox{\centering \footnotesize \hspace{5pt}Camera Motion Blur}}
  \hfill
  \begin{subfigure}{0.3198\linewidth}
    \includegraphics[width=\linewidth]{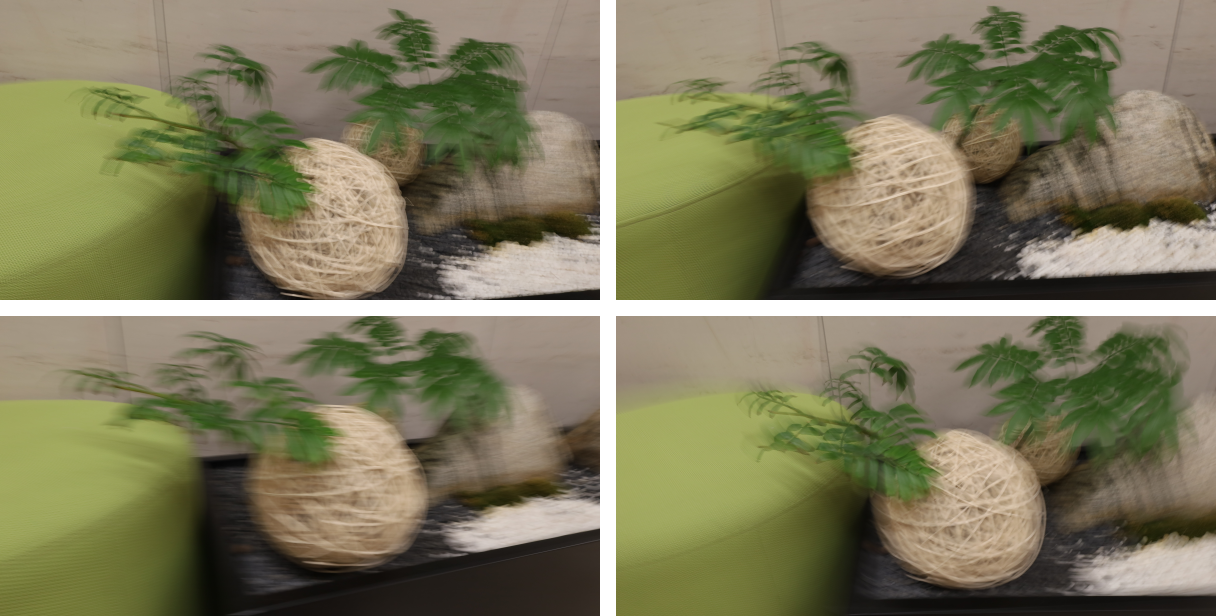}
    \label{subfig:teaser_blur_input}
  \end{subfigure}
  \hfill
  \begin{subfigure}{0.324\linewidth}
    \includegraphics[width=\linewidth]{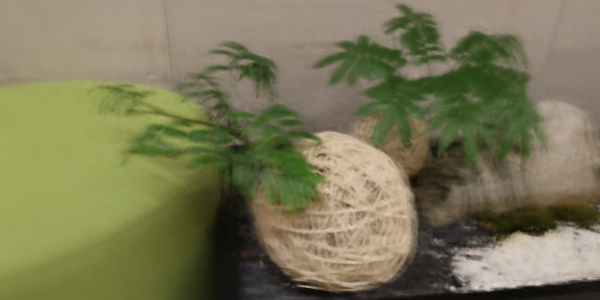}
    \label{subfig:teaser_blur_naive}
  \end{subfigure}
  \hfill
  \begin{subfigure}{0.324\linewidth}
    \includegraphics[width=\linewidth]{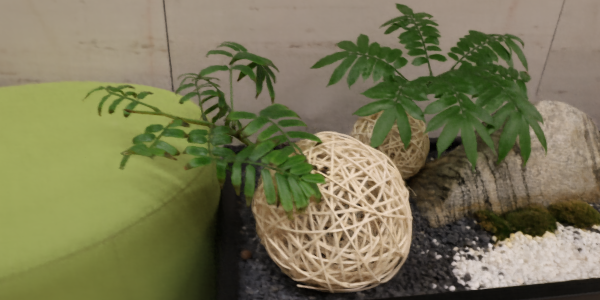}
    \label{subfig:teaser_blur_ours}
  \end{subfigure}
  \hfill
  
  \vspace{\figtofig}
  \rotatebox[origin=c]{90}{\makebox{\centering \footnotesize \hspace{10pt}Defocus Blur}}
  \hfill
  \begin{subfigure}{0.3198\linewidth}
    \includegraphics[width=\linewidth]{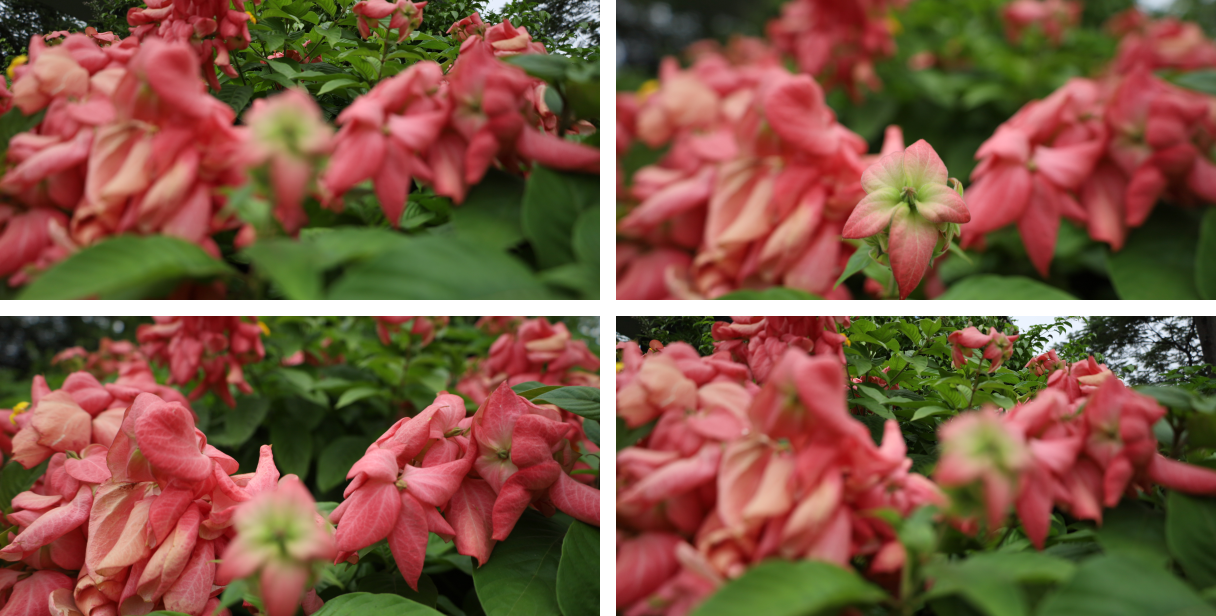}
    \label{subfig:teaser_defocus_input}
    \vspace{\figtosubcap}
    \caption{Samples of Blurry Source Views}
  \end{subfigure}
  \hfill
  \begin{subfigure}{0.324\linewidth}
    \includegraphics[width=\linewidth]{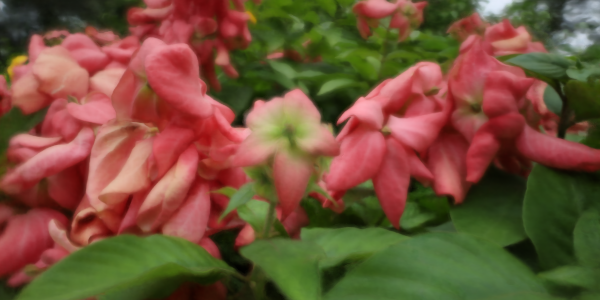}
    \label{subfig:teaser_defocus_naive}
    \vspace{\figtosubcap}
    \caption{Novel Views from NeRF}
  \end{subfigure}
  \hfill
  \begin{subfigure}{0.324\linewidth}
    \includegraphics[width=\linewidth]{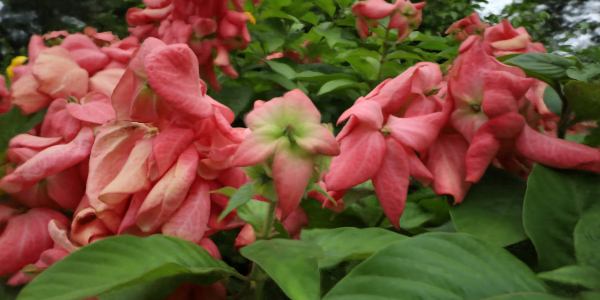}
    \label{subfig:teaser_defocus_ours}
    \vspace{\figtosubcap}
    \caption{Novel Views from Deblur-NeRF}
  \end{subfigure}
  \hfill
  \vspace{-0.2cm}
  \caption{Given a set of blurry multi-view input images (a), the original NeRF implementation reconstructs blurry novel views (b). Our method is able to recover a sharp radiance field and synthesize clear novel views (c). Our proposed approach can handle both camera motion blur (first row) and defocus blur (second row). Please refer to the \textit{supplementary material} for video results.}
  \label{fig:teaser}
\end{center}%
}]

\maketitle

\footnotetext[1]{Author did this work during the internship at Tencent AI Lab.}
\begin{abstract}
Neural Radiance Field (NeRF) has gained considerable attention recently for 3D scene reconstruction and novel view synthesis due to its remarkable synthesis quality. However, image blurriness caused by defocus or motion, which often occurs when capturing scenes in the wild, significantly degrades its reconstruction quality. To address this problem, We propose Deblur-NeRF, the first method that can recover a sharp NeRF from blurry input. We adopt an analysis-by-synthesis approach that reconstructs blurry views by simulating the blurring process, thus making NeRF robust to blurry inputs. The core of this simulation is a novel Deformable Sparse Kernel (DSK) module that models spatially-varying blur kernels by deforming a canonical sparse kernel at each spatial location. The ray origin of each kernel point is jointly optimized, inspired by the physical blurring process. This module is parameterized as an MLP that has the ability to be generalized to various blur types. Jointly optimizing the NeRF and the DSK module allows us to restore a sharp NeRF. We demonstrate that our method can be used on both camera motion blur and defocus blur: the two most common types of blur in real scenes. Evaluation results on both synthetic and real-world data show that our method outperforms several baselines. The synthetic and real datasets along with the source code is publicly available at {\small \url{https://limacv.github.io/deblurnerf/}}.  
\end{abstract}

\vspace{-0.5cm}
\section{Introduction}
\label{sec:intro}

Tremendous progress has been witnessed in the past few years in novel view synthesis, where an intermediate 3D representation is reconstructed from sparse input views to interpolate or extrapolate arbitrary novel views. Recently, NeRF~\cite{NeRF} emerged as an effective scene representation that achieves photorealistic rendering results. It models a static scene as a continuous volumetric function that maps 3D location and 2D view direction to color and density. This function is parameterized as a multilayer perceptron (MLP), and its output can be rendered by volume rendering techniques in a differentiable manner.

To reconstruct a NeRF, several images from different views are needed. While the original approach for training NeRF works well when these images are well captured and calibrated, it would produce obvious artifacts when blur occurs. For example, when using a long exposure setting to capture a low-light scene, the images are more sensitive to camera shake, resulting in camera motion blur. Furthermore, defocus blur is inevitable when capturing scenes with large depth variation using a large aperture. These blurs will significantly decrease the quality of the reconstructed NeRF, resulting in artifacts in the rendered novel views.

Many works have recently been proposed to tackle abnormal input while training NeRF. NeRF-W~\cite{NeRF_misc_nerf_w} focuses on images with illumination change and moving objects. Mip-NeRF~\cite{NeRF_misc_mipnerf} improves the NeRF when the input spans different scales. Distortion in input is considered and calibrated simultaneously in SCNeRF \cite{NeRF_calib_SCNeRF}. To the best of our knowledge, none has considered addressing the problem of training NeRF from blurry input images. One solution is to first deblur the input in image space and then train the NeRF with deblurred images, which we refer to as the image-space baseline. This baseline improves the novel view synthesis quality of NeRF to some extend by utilizing recent single image or video deblurring methods. However, the single-image deblur methods fail to aggregate information from neighbor views and can not guarantee the multi-view consistent result. Video-based methods manage to take multi-frame into consideration, usually relying on image space operations such as optical flows and feature correlation volumes. However, these methods fail to exploit the 3D geometry of the scene, leading to inaccurate correspondences across views, especially when they have a large baseline. On the contrary, our method deblurs by aggregating information from all observations with full awareness of the 3D scene. \ignore{To summarize, there are no existing methods or direct solutions that consider multi-view consistency to reconstruct a sharp NeRF with blurry input, which will be explored in this work.}

In this paper, we propose Deblur-NeRF, an effective framework that explicitly models the blurring process in the network, and is capable of restoring a sharp NeRF from blurry input. We model the blurring process by convolving a clean image using a blur kernel similar to blind deconvolution methods \cite{survey_deconv}. A novel deformable sparse kernel (DSK) module is proposed to model the blur kernel inspired by the following observations. First, convolving with dense kernels is infeasible for scene representations such as NeRF due to the dramatic increase in computation and memory usage during rendering. To address this, DSK uses sparse rays to approximate the dense kernel. Second, we show that the actual blurring process involves combining rays from different origins, which motivates us to jointly optimize the ray origins. Finally, to model spatially-varying blur kernels, we deform a canonical sparse kernel at each 2D spatial location.
\ignore{Our method is inspired by blind deconvolution \cite{survey_deconv} that models the blur as convolving a clean image using a blur kernel. However, classical deconvolution methods cannot be directly applied to scene representations like NeRF due to the dramatic increase in computation and memory usage during rendering. To address this, we propose a deformable sparse kernel (DSK) module that is optimized to approximate the dense kernel to reduce the rendering overhead. To model spatially-varying blur kernels in an image and simplify the learning process, we deform a canonical sparse kernel at each spatial location.} 
The deformation is parameterized as an MLP that can be generalized to different types of blur. During training, we jointly optimize the DSK and a sharp NeRF with only blurry input as supervision, while in the inference stage, clear novel views can be rendered by removing the DSK. We conduct extensive experiments on both synthetic and real datasets with two types of blur: camera motion blur and defocus blur. Results show that the proposed method outperforms the original NeRF and image-space baselines (i.e., combining NeRF with the state-of-the-art image or video deblurring methods), for these two blur types, as shown in \cref{fig:teaser} and the experiments section. Our contributions can be summarized as follows:
\begin{itemize}
    \vspace{-0.1cm}
    \item We propose the first framework that can reconstruct a sharp NeRF from blurry input.
    \vspace{-0.1cm}
    \item We propose a deformable sparse kernel module that enables us to effectively model the blurring process and is generalizable for different types of blur.
    \vspace{-0.1cm}
    \item We analyze the physical blurring process and extend the 2D kernel to 3D space by considering the translation of the ray origin for each kernel point.
    
\end{itemize}

\section{Related Work}

\tinysection{Neural radiance field.}
Our work extends the NeRF \cite{NeRF}, a coordinate-based implicit 3D scene representation, which has gained popularity over the past few years due to its state-of-the-art novel view synthesis results. The success of NeRF has inspired many follow-up works that extend the NeRF \cite{NeRF_dyn_dnerf,NeRF_dyn_hypernerf,NeRF_dyn_nerfies,NeRF_dyn_neuralsceneflow,NeRF_fast_baking,NeRF_fast_kilonerf,NeRF_GAN_graf,NeRF_GAN_giraffe,NeRF_hdr}. \ignore{to dynamic scenes \cite{NeRF_dyn_dnerf,NeRF_dyn_hypernerf,NeRF_dyn_nerfies,NeRF_dyn_neuralsceneflow}, improve the render speed \cite{NeRF_fast_baking,NeRF_fast_kilonerf}, introduce generative model to NeRF \cite{NeRF_GAN_graf,NeRF_GAN_giraffe}, etc.} Several works have explored to train the NeRF with non-ideal input. For example, BRAF \cite{NeRF_calib_braf}, NeRF$--$ \cite{NeRF_calib_nerf__} and GNeRF \cite{NeRF_calib_GNeRF} try to train the NeRF without camera poses. SCNeRF \cite{NeRF_calib_SCNeRF} focuses on jointly calibrating a more complex non-linear camera model. To address the NeRF training under uncontrolled, in-the-wild photographs, NeRF-W \cite{NeRF_misc_nerf_w} introduces several extensions to NeRF that successfully model the inconsistent appearance variations and transient objects across views. PixelNeRF \cite{NeRF_misc_pixelnerf} reconstructs a neural volume with only one or few images. Moreover, Jonathan \etal proposes Mip-NeRF \cite{NeRF_misc_mipnerf} which improves the NeRF under input with different scales, producing anti-aliased results. However, training NeRF with blurry images is still an unexplored area, as none of the aforementioned works seem to explicitly consider this kind of degradation. 

\tinysection{Single image deblurring.}
\ignore{*already said multiple times:* Image blur is a common artifact that usually happens when the scene or camera moves during exposure or the scene objects lie out of the focus plane, which is called camera motion blur and defocus blur, respectively.} Image deblurring aims at recovering a sharp image from blurry input. Usually, the blurry image is modeled as the sharp image convolved with a blur kernel, and the deblurring process is formulated as jointly solving the sharp image and the kernel given the blurry observation. This task is ill-posed since there are many sets of image-blur pairs that can synthesize the observed blurry image~\cite{survey_deblurring}. Classical blind deblur algorithms tackle the ill-posedness by introducing hand-crafted or learned image priors while optimizing the sharp image and kernel, such as total variation \cite{Blur_prior_TV,Blur_prior_TVDeconv}, normalized gradient sparsity \cite{Blur_prior_sparsity} and unnatural $l_0$ \cite{Blur_prior_unnatural}. Since blur in real world photographs is usually spatially-varying, many works try to reparameterize the blur kernels to a smaller solve space. Early work uses projective motion blur \cite{Blur_SV_homo} which fits spatially-varying blur kernels using multiple homography, while region-based methods assume piece-wise constant \cite{Blur_SV_regionConst} or piece-wise projective \cite{Blur_SV_bilayer}. Moreover, a depth-based model is used to optimize the depth map and camera poses jointly \cite{Blur_SV_poseanddepth,Blur_SV_worldblur}. Another approach to model blur kernels is to use optical flow \cite{Blur_SV_generalizedvideo}. These methods either make a strong assumption on the blur pattern, or can only model one specific type of blur. In contrast, our method models the spatially-varying kernel using MLP, which can be generalized to different blur types. The recent trend of image deblurring is to introduce deep neural networks that directly map the blurry image to the latent sharp image \cite{Blur_CNN_1stlearn,Blur_CNN_2ndlearn,Blur_CNN_GAN1,Blur_CNN_GAN2,Blur_CNN_inthewild,Blur_CNN_multiscale,Blur_CNN_scalerecurrent,Blur_CNN_MPR,Defocus_IFAN,Defocus_KPAC}. These approaches have outperformed traditional methods. However, this line of work highly depends on the training data and the methods often have difficulty generalizing to unseen blur types in the real world \cite{Blur_EBKS}.

\tinysection{Multi-image deblurring.}
Deblurring with multi-image settings poses new challenges in aggregating information across frames and preserving temporal consistency. Optical flow is a useful tool for registering neighbor frames to the reference frame \cite{Blur_SV_generalizedvideo,Blur_video_cascaded}. However, estimating accurate optical flow is difficult and ill-posed, especially when the input is blurry. With the advancement of deep learning, one can design flow-free methods by concatenating multiple frames and directly restoring the clean frame using a CNN \cite{Blur_video_DVD}. Another option is to use recurrent structure that propagates features across frames \cite{Blur_video_dynamictemp,Blur_video_intraframe,Blur_video_strecurrent,Blur_video_PVD}. Li \etal \cite{Blur_video_arvo} extends the optical flow to feature correlation volume, which greatly improves the performance. Similarly, Son \etal \cite{Blur_video_PVD} propose pixel volume that relaxes the requirement for accurate flow. However, these multi-image deblurring methods, which are built on image space operations, fail to exploit the 3D geometry of the scene and have difficulty addressing the multi-view input with a large baseline.

\vspace{-0.1cm}
\section{Preliminary}
We first review the NeRF representation \cite{NeRF} for a static 3D scene. A NeRF defines the scene as a continuous volumetric function that maps a 3D position $\mathbf{x}$ and a 2D view direction $\mathbf{d}$ to color $\mathbf{c}$ and volume density $\sigma$. Formally:
\begin{equation}
    (\mathbf{c}, \sigma) = F_\mathbf{\Theta}\left(
    \gamma_{L_x}(\mathbf{x}), \gamma_{L_d}(\mathbf{d})
    \right),
\end{equation}
where $F_\mathbf{\Theta}$ represents an MLP with parameters $\mathbf{\Theta}$, and $\gamma_L(\cdot)$ is the positional encoding that maps each element of a vector into a higher dimensional frequency space:
\begin{equation}
    \gamma_L(x)=\left[\sin{\pi x},\cos{\pi x},...,\sin{2^{L-1}\pi x},\cos{2^{L-1}\pi x}\right]^\text{T},
\end{equation}
where the hyper-parameter $L$ indicates the highest frequency used in the mapping and can be used to control the smoothness of the scene function \cite{misc_PE}. To render a pixel centered at image coordinate $\mathbf{p}$, we first emit a ray $\mathbf{r_p}(t) = \mathbf{o} + t\mathbf{d_p}$ from camera projection center $\mathbf{o}$ along the viewing direction $\mathbf{d_p}$.\ignore{The subscript $\mathbf{p}$ indicates that $\mathbf{d_p}$ is determined by the pixel location $\mathbf{p}$.} Then a sampling strategy is used to determine $D$ sorted distances $\{t^{(i)}\}_{i=1}^D$ between predefined near and far planes $t^{(0)}$ and $t^{(D+1)}$. We estimate the color $\mathbf{c}^{(i)}$ and density $\sigma^{(i)}$ at each sample point $\mathbf{r_p}(t^{(i)})$ using $F_\mathbf{\Theta}$. The final color of the pixel is computed as:
\begin{equation}
    \hat{\mathbf{c}}_\mathbf{p} = \hat{\mathbf{c}}(\mathbf{r}_\mathbf{p}) = \sum_{i=1}^D T^{(i)} \left(1 - \exp(-\sigma^{(i)} \delta^{(i)})\right) \mathbf{c}^{(i)} ,
    \label{eq:nerfrender}
\end{equation}
where $\delta^{(i)} = t^{(i+1)}-t^{(i)}$ is the distance between adjacent samples, and $T^{(i)} = \exp(-\sum_{j=1}^{i-1}\sigma^{(j)}\delta^{(j)})$. In this paper we use $\mathbf{c_p}$ and $\mathbf{c}(\mathbf{r_p})$ interchangeably. Note that this rendering process is trivially differentiable. 


\begin{figure*}[t]
\centering
\includegraphics[width=0.98\linewidth]{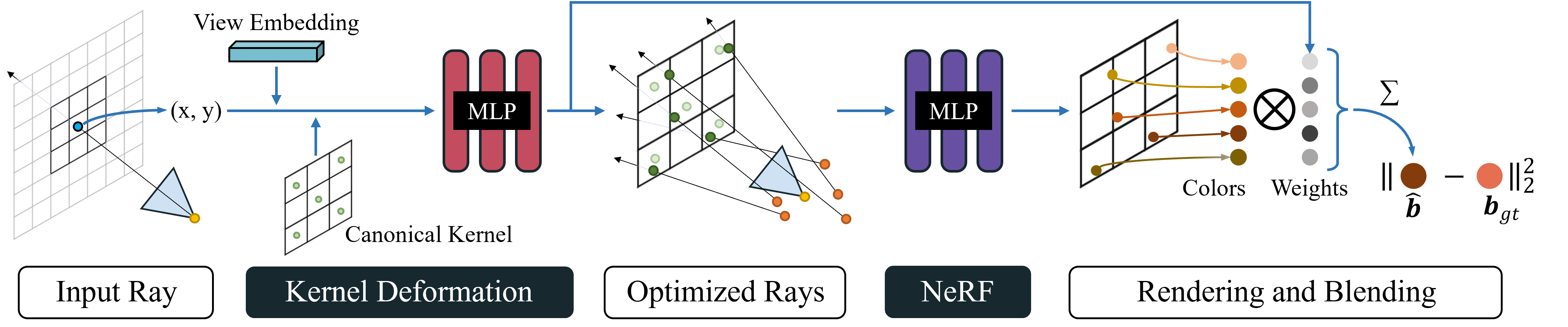}
\vspace{-0.2cm}
\caption{An overview of our training framework. When rendering a ray, we first predict $N$ sparse optimized rays based on a canonical kernel along with their weights. After rendering these rays, we combine the results to get the blurry pixel $\mathbf{\hat{b}}$. Note that when testing, we can directly render the rays without kernel deformation resulting in a sharp image.}
\vspace{-8pt}
\label{fig:pipeline}
\end{figure*}

\section{Method}
Our task is to train the NeRF with blurry input. The training pipeline is visualized in \cref{fig:pipeline}. The core idea is to explicitly model the blurring process and seek to optimize a sharp NeRF and the blur parameters jointly so that the synthesized blurry images match the input. Specifically, to render a blurry pixel during training, we first generate multiple optimized rays using a newly proposed Deformable Sparse Kernel (DSK) module, which mimics the blurring process. We render these rays using the NeRF, and blend results to get the finally blurry color, which is then supervised by the blurry input. Note that in the inference stage, we can directly render the NeRF without the DSK to get sharp novel views. We describe the DSK as well as some other designs in the following subsections.
\vspace{-0.1cm}
\subsection{Deformable Sparse Kernel}
Akin to most image deblurring algorithms \cite{survey_deblurring,survey_deconv}, we model the blurring process by convolving the sharp image with a blur kernel $h$:
\begin{equation}
    \mathbf{b}_\mathbf{p} = \mathbf{c}_\mathbf{p} * h,
    \label{eq:blurconv0}
\end{equation}
where $\mathbf{c}_\mathbf{p}$ is the color of sharp pixel at $\mathbf{p}$, which ideally is also the output of a sharp NeRF in our model. $\mathbf{b}_\mathbf{p}$ is the corresponding blurry color and $*$ stands for the convolution operator.\ignore{A more concrete model should also include the noise, which we omit 
because we find that the NeRF can already removes the signal-uncorrelated noise.} The support of the blur kernel is usually defined in a $K \times K$ window centered at $\mathbf{p}$. To compute $\mathbf{b}_\mathbf{p}$, we take the sum of element-wise multiplication of $\mathbf{c}_\mathbf{p}$ and $h$ inside the window. This can be computed efficiently in the map-based image representation. However, a problem arises when $\mathbf{c}_\mathbf{p}$ is modeled as a NeRF because rendering becomes quite computation and memory consuming. For each pixel, there are $K \times K$ rays that need to be rendered in the supporting window, thus making training infeasible. Therefore, we propose to approximate the dense blur kernel with a small number of sparse points:
\begin{equation}
    \mathbf{b}_\mathbf{p} = \sum_{\mathbf{q}\in \mathcal{N}(\mathbf{p})} w_\mathbf{q} \mathbf{c}_\mathbf{q}\text{, w.r.t.} \sum_{\mathbf{q}\in \mathcal{N}(\mathbf{p})} w_\mathbf{q} = 1\text{,}
\end{equation}
where $\mathcal{N}(\mathbf{p})$ is the set of $N$ locations sparsely distributed around $\mathbf{p}$ that compose the support of our sparse kernel. $w_\mathbf{q}$ is the corresponding weight at each location. We set $N$ to be a fixed number and ablate this hyper-parameter in our experiments (\cref{sec:experiments}). 
Note that $\mathbf{q}$ is a continuous value and we can jointly optimize the locations $\mathcal{N}(\mathbf{p})$, $w_\mathbf{q}$ and the NeRF so that the best sparse kernel is regressed. 

The blur kernel is usually spatially-varying in real world images. Inspired by the NeRF that uses an MLP as a continuous 5D function, we also choose to use an MLP to model the spatially-varying kernel\ignore{ from a 2D pixel coordinate}. Specifically, for each input view, we assign ``canonical kernel locations'' $\mathcal{N}'(\mathbf{p}) = \{\mathbf{q}'_i\}_{i=0}^{N-1}$, and use an MLP to deform the locations and while also predicting the weights:
\begin{equation}
    (\Delta \mathbf{q}, w_\mathbf{q}) = G_\mathbf{\Phi}(\mathbf{p}, \mathbf{q}', \mathbf{l}), \text{where } \mathbf{q}' \in \mathcal{N}'(\mathbf{p}).
    \label{eq:kernelmlp}
\end{equation}
Here $G_\mathbf{\Phi}$ indicates an MLP with parameter $\mathbf{\Phi}$ and $\mathbf{l}$ is a learned view embedding. This view embedding is necessary since the blur patterns usually differ across views. Optimizing a different view embedding for each view allows the DSK module to fit a different blur kernel for each view. In our experiments we set the view embedding to be a vector of length 32. We compute the final sparse kernel location in $\mathcal{N}(\mathbf{p})$ as $\mathbf{q} = \mathbf{q}' + \Delta \mathbf{q}$. Note that we need to forward the MLP $G_\mathbf{\Phi}$ for $N$ times to get all the deformed locations. One option that may potentially boost the performance is introducing positional encoding to $G_\mathbf{\Phi}$ by replacing the input $\mathbf{p}$ in \cref{eq:kernelmlp} with $\gamma(\mathbf{p})$. However, we find this operation does not help to improve the quality. One possible reason is that the spatially-varying kernel changes gradually along spatial positions without high-frequency variation.
\vspace{-0.1cm}
\subsection{Convolution with Irradiance}
As pointed out by Chen \etal \cite{misc_DeblurCRF}, this blur convolution model should be applied to scene irradiance instead of image intensity. A more physically correct model should be $\mathbf{b}_p = f(\mathbf{c}'_p * h)$, where $\mathbf{c}'$ indicates the scene irradiance and $f(\cdot)$ is the camera response function (CRF) that maps the scene irradiance to image intensity. A nonlinear CRF will increase the complexity of the blur kernel and make the learning of DSK difficult if the linear model in \cref{eq:blurconv0} is used, especially in high contrast regions \cite{survey_deblurring}. To compensate for the nonlinear CRF, we assume that our sharp NeRF predicts colors in linear space and adopt a simple gamma correction function in the final output:
\begin{equation}
    \mathbf{b}_\mathbf{p} = g(\sum_{\mathbf{q}\in \mathcal{N}(\mathbf{p})} w_\mathbf{q} \mathbf{c}'_\mathbf{q}),
    \label{eq:gamma+spasekernel}
\end{equation}
where $g(\mathbf{c}') = \mathbf{c}'^{\frac{1}{2.2}}$ is the gamma correction function. More complex CRFs could be used to model real world cameras, such as pre-calibrated CRFs, or jointly optimizing the CRFs during training. But we find this simple scheme is enough to compensate the nonlinearity in the imaging process and improve the quality. More discussions about modeling the CRF can be found in \textit{supplementary material}.

\begin{figure}
    \centering
    \includegraphics[width=\linewidth]{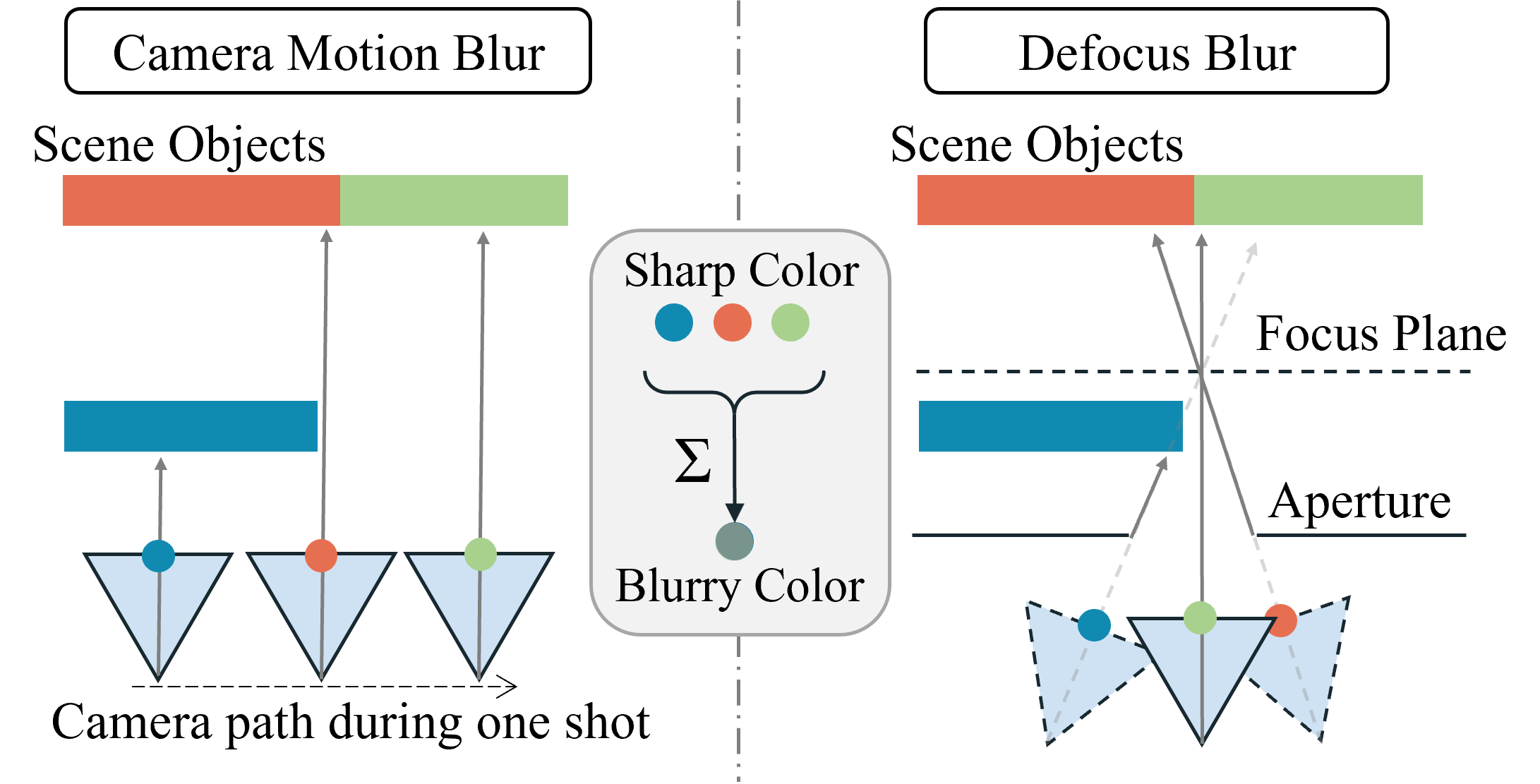}
    \vspace{-20pt}
    \caption{Top-down view of the blurring processes caused by camera motion and defocus. The blurry color is blended from all the rays destined at a pixel. In defocus blur, the rays of a pixel are scattered to different directions at the focus plane, which is equivalent to a mixture of rays being emitted from different camera centers.}
    \label{fig:trans_necessity}
    \vspace{-0.2cm}
\end{figure}

\subsection{Optimizing the Ray Origin}
One observation of the convolutional model is that it is a 2D approximation of the actual blur model. In the convolutional model, the blurry result is a combination of neighbor pixels, which are the rendering results of neighbor rays with the same camera center as the origin. However, the actual blurring process usually involves blending rays cast from different origins. Consider the two blurring processes shown in \cref{fig:trans_necessity}. When capturing camera motion blur, the camera center moves during one shot, leading to the change of the ray origins. And for defocus blur, the ray gets scattered to different directions which is equivalent to a mixture of rays from different origins. When the scene is mostly planar, the translation of the ray origin can be well approximated by the 2D translation of the pixel position. However, due to the parallax effect and occlusion, this is not the case when there is depth discontinuity. Since we have access to the 3D scene representation, we can develop kernels that consider the change of ray origins. Thus we jointly optimize the translation of the ray origin of each sparse kernel location. Specifically, we jointly predict the origin translation for each kernel location as follows \cref{eq:kernelmlp}:
\begin{equation}
    (\Delta \mathbf{o_q}, \Delta \mathbf{q}, w_\mathbf{q}) = G_\mathbf{\Phi}(\mathbf{p}, \mathbf{q}', \mathbf{l}) \text{, } \mathbf{q}' \in \mathcal{N}'(\mathbf{p}),
    \label{eq:kernelmlpfinal}
\end{equation}
and then generate rays by:
\begin{equation}
    \mathbf{r_q} = (\mathbf{o} + \Delta \mathbf{o}_{\mathbf{q}}) + t \mathbf{d}_{\mathbf{q}} \text{,  } \mathbf{q} = \mathbf{q}'+\Delta \mathbf{q}.
    \label{eq:generaterayfinal}
\end{equation}
These optimized rays are rendered and combined to get the final blurry pixels. 

The training process is summarized as follows: we first predict tuples $\{\Delta \mathbf{o_q}, \Delta \mathbf{q}, w_\mathbf{q}\}_{q\in \mathcal{N}(p)}$ using \cref{eq:kernelmlpfinal}, with which we generate multiple optimized rays $\{\mathbf{r_q}\}_{q\in \mathcal{N}(p)}$ by deforming the canonical sampling location and optimizing the ray origin as in \cref{eq:generaterayfinal}. We render these rays to get $\mathbf{c}'_q$ using \cref{eq:nerfrender} and blend to get a blurry pixel $\mathbf{\hat{b}}_p$ using \cref{eq:gamma+spasekernel}. This synthetic blurry pixel is supervised by the corresponding ground truth pixel color $\mathbf{b}_{gt}$:
\begin{equation}
    \mathcal{L}_{reconstruct} = \sum_{\mathbf{p}\in \mathcal{R}}\|\mathbf{\hat{b}}_p - \mathbf{b}_{gt}\|_2^2,
    \label{eq:reconstructloss}
\end{equation}
where $\mathcal{R}$ is the set of pixels in each batch. Note that our pipeline is only used during training. At test time, we can directly render the sharp results using the restored sharp NeRF with gamma correction.

\subsection{Aligning the NeRF}
As shown in the experiments, if we freely optimize all the learnable components, i.e., the NeRF and the deformable sparse kernel, the reconstructed NeRF may undergo some non-rigid distortion. This is in line with expectations because it is possible that the scene represented by the NeRF and the learned kernel deform together without affecting the reconstructed blurry result. However this is usually not desired. To constrain the NeRF model to align with the observations, we first initialize the deformable sparse kernel so that all the optimized rays $\mathbf{r_q}$ are close to the input ray $\mathbf{r_p}$. 
This is implemented by multiplying a small gain of $\epsilon = 0.1$ to each element of the output tuples $(\Delta \mathbf{o_q}, \Delta \mathbf{q}, w_\mathbf{q})$. As a result, the ray origins $\mathbf{o_q}$ and the kernel points $\mathbf{q}$ are initialized to be close to the camera centers and the pixel locations, respectively. And all kernel points will have roughly the same weights at the beginning of training.
We additionally introduce an alignment loss that forces one of the optimized rays $\mathbf{r_q}$ to be similar to the input ray $\mathbf{r_p}$: 
\begin{equation}
    \mathcal{L}_{align} = \|\mathbf{q}_0 - \mathbf{p}\|_2 + \lambda_o \|\Delta \mathbf{o}_{\mathbf{q}_0}\|_2,
\end{equation}
where $\mathbf{q}_0$ is a fixed element in $\mathcal{N}(\mathbf{q})$. By applying $\mathcal{L}_{align}$, we supervise $\mathbf{q}_0$ to be the center of the kernel.
We set $\lambda_o = 10$ in all of our experiments.

Our final loss is a combination of the NeRF reconstruction loss and the alignment loss:
\begin{equation}
    \mathcal{L} = \mathcal{L}_{reconstruct} + \lambda_a \mathcal{L}_{align}.
\end{equation}
In our experiments we set $\lambda_a = 0.1$.

\section{Experiments}
\label{sec:experiments}
\subsection{Implementation Details}
\tinysection{Training.}  
We build our deformable sparse kernel on the Pytorch re-implementation of the NeRF \cite{misc_nerfpytorch}. We use a batch size of 1024 rays, each sample at $64$ coordinates in the coarse volume and $64$ additional coordinates in the fine volume. We set the number of sparse locations $N = 5$. We use the Adam optimizer \cite{misc_adam} with default parameters. We schedule the learning rate to start at $5\times 10^{-4}$ and decay exponentially to $8\times 10^{-5}$ over the coarse of the optimization. We train each scene for 200k iterations on a single NVIDIA V100 GPU. We adopt the same MLP structure of $F_\mathbf{\Theta}$ as the original NeRF \cite{NeRF}, and for $G_\mathbf{\Phi}$, we use MLP with $4$ fully-connected hidden layers, each layer having $64$ channels and ReLU activations. We also add a shortcut that connects the first layer to the last layer.

\begin{table*}
\centering
\footnotesize
\setlength{\tabcolsep}{1pt}
\newcolumntype{Y}{>{\centering\arraybackslash}X}
\begin{tabularx}{0.999\linewidth}{l||YYY|YYY|YYY|YYY|YYY|YYY}
\toprule
  & \multicolumn{3}{c}{\scshape Factory}
  & \multicolumn{3}{c}{\scshape Cozyroom}
  & \multicolumn{3}{c}{\scshape Pool}
  & \multicolumn{3}{c}{\scshape Tanabata} 
  & \multicolumn{3}{c}{\scshape Trolley}
  & \multicolumn{3}{c}{\scshape Average}  \\
\textsl{Camera Motion} & \multicolumn{1}{c}{\scriptsize PSNR$\uparrow$} & \multicolumn{1}{c}{\scriptsize SSIM$\uparrow$} & \multicolumn{1}{c}{\scriptsize LPIPS$\downarrow$} & \multicolumn{1}{c}{\scriptsize PSNR$\uparrow$} & \multicolumn{1}{c}{\scriptsize SSIM$\uparrow$} & \multicolumn{1}{c}{\scriptsize LPIPS$\downarrow$} & \multicolumn{1}{c}{\scriptsize PSNR$\uparrow$} & \multicolumn{1}{c}{\scriptsize SSIM$\uparrow$} & \multicolumn{1}{c}{\scriptsize LPIPS$\downarrow$} & \multicolumn{1}{c}{\scriptsize PSNR$\uparrow$} & \multicolumn{1}{c}{\scriptsize SSIM$\uparrow$} & \multicolumn{1}{c}{\scriptsize LPIPS$\downarrow$} & \multicolumn{1}{c}{\scriptsize PSNR$\uparrow$} & \multicolumn{1}{c}{\scriptsize SSIM$\uparrow$} & \multicolumn{1}{c}{\scriptsize LPIPS$\downarrow$} & \multicolumn{1}{c}{\scriptsize PSNR$\uparrow$} & \multicolumn{1}{c}{\scriptsize SSIM$\uparrow$} & \multicolumn{1}{c}{\scriptsize LPIPS$\downarrow$}  \\
\midrule
w/o gamma   & 23.27  & .6908  & .3210   & \cellcolor{third}29.86  & \cellcolor{third}.8964  & \cellcolor{third}.0564   & \cellcolor{third}31.29  & \cellcolor{third}.8635  & \cellcolor{third}.1339   & \cellcolor{third}25.50  & \cellcolor{third}.8211  & .1472   & \cellcolor{third}25.44  & \cellcolor{third}.8071  & \cellcolor{third}.1513   & \cellcolor{third}27.07  & \cellcolor{third}.8158  & .1620    \\
w/o align  & \cellcolor{third}24.95  & \cellcolor{third}.7419  & \cellcolor{second}.2780   & 25.53  & .8032  & .0636   & 27.45  & .7389  & .1443   & 24.77  & .8086  & \cellcolor{second}.1282   & 24.71  & .8036  & \cellcolor{best}.1324   & 25.48  & .7792  & \cellcolor{third}.1493    \\
w/o origin opt.  & \cellcolor{second}25.29  & \cellcolor{second}.7657  & \cellcolor{third}.2827   & \cellcolor{second}31.86  & \cellcolor{second}.9244  & \cellcolor{second}.0479   & \cellcolor{best}31.64  & \cellcolor{best}.8691  & \cellcolor{best}.1216   & \cellcolor{second}26.20  & \cellcolor{second}.8475  & \cellcolor{third}.1523   & \cellcolor{second}25.53  & \cellcolor{second}.8199  & .1774   & \cellcolor{second}28.11  & \cellcolor{second}.8453  & \cellcolor{second}.1564    \\
Ours          & \cellcolor{best}25.60  & \cellcolor{best}.7750  & \cellcolor{best}.2687 & \cellcolor{best}32.08 & \cellcolor{best}.9261 & \cellcolor{best}.0447  & \cellcolor{second}31.61  & \cellcolor{second}.8682  & \cellcolor{second}.1246 & \cellcolor{best}27.11 & \cellcolor{best}.8640 & \cellcolor{best}.1228  & \cellcolor{best}27.45 & \cellcolor{best}.8632 & \cellcolor{second}.1363  & \cellcolor{best}28.77 & \cellcolor{best}.8593 & \cellcolor{best}.1400   \\
\midrule
\midrule
\textsl{Defocus}
              & \multicolumn{1}{c}{\scriptsize PSNR$\uparrow$} & \multicolumn{1}{c}{\scriptsize SSIM$\uparrow$} & \multicolumn{1}{c}{\scriptsize LPIPS$\downarrow$} & \multicolumn{1}{c}{\scriptsize PSNR$\uparrow$} & \multicolumn{1}{c}{\scriptsize SSIM$\uparrow$} & \multicolumn{1}{c}{\scriptsize LPIPS$\downarrow$} & \multicolumn{1}{c}{\scriptsize PSNR$\uparrow$} & \multicolumn{1}{c}{\scriptsize SSIM$\uparrow$} & \multicolumn{1}{c}{\scriptsize LPIPS$\downarrow$} & \multicolumn{1}{c}{\scriptsize PSNR$\uparrow$} & \multicolumn{1}{c}{\scriptsize SSIM$\uparrow$} & \multicolumn{1}{c}{\scriptsize LPIPS$\downarrow$} & \multicolumn{1}{c}{\scriptsize PSNR$\uparrow$} & \multicolumn{1}{c}{\scriptsize SSIM$\uparrow$} & \multicolumn{1}{c}{\scriptsize LPIPS$\downarrow$} & \multicolumn{1}{c}{\scriptsize PSNR$\uparrow$} & \multicolumn{1}{c}{\scriptsize SSIM$\uparrow$} & \multicolumn{1}{c}{\scriptsize LPIPS$\downarrow$}  \\
\midrule
w/o gamma   & 25.92  & \cellcolor{third}.8170  & .1579   & \cellcolor{second}31.18  & \cellcolor{third}.9078  & .0556   & \cellcolor{third}29.97  & \cellcolor{third}.8095  & .2082   & \cellcolor{third}25.00  & \cellcolor{third}.8172  & .1299   & 24.37  & .7819  & .1680   & \cellcolor{third}27.29  & \cellcolor{third}.8267  & .1439    \\
w/o align  & \cellcolor{third}26.31  & .8051  & \cellcolor{third}.1493   & \cellcolor{third}27.76  & .8583  & \cellcolor{third}.0545   & 28.00  & .7481  & \cellcolor{third}.2078   & 24.86  & .8149  & \cellcolor{second}.1021   & \cellcolor{third}24.50  & \cellcolor{third}.7977  & \cellcolor{best}.1292   & 26.28  & .8048  & \cellcolor{second}.1286   \\
w/o origin opt.  & \cellcolor{second}28.00  & \cellcolor{second}.8584  & \cellcolor{second}.1344   & \cellcolor{best}31.85  & \cellcolor{second}.9173  & \cellcolor{second}.0506   & \cellcolor{second}30.21  & \cellcolor{second}.8172  & \cellcolor{second}.2023   & \cellcolor{second}25.75  & \cellcolor{second}.8444  & \cellcolor{third}.1104   & \cellcolor{second}24.82  & \cellcolor{second}.8014  & \cellcolor{third}.1590   & \cellcolor{second}28.13  & \cellcolor{second}.8477  & \cellcolor{third}.1313    \\
Ours          & \cellcolor{best}28.03  & \cellcolor{best}.8628  & \cellcolor{best}.1127   & \cellcolor{best}31.85  & \cellcolor{best}.9175  & \cellcolor{best}.0481   & \cellcolor{best}30.52  & \cellcolor{best}.8246  & \cellcolor{best}.1901   & \cellcolor{best}26.25  & \cellcolor{best}.8517  & \cellcolor{best}.0995   & \cellcolor{best}25.18  & \cellcolor{best}.8067  & \cellcolor{second}.1436   & \cellcolor{best}28.37  & \cellcolor{best}.8527  & \cellcolor{best}.1188    \\
\bottomrule
\end{tabularx}
\vspace{\figtocapexp}
\caption{Ablations of our method in the synthetic scenes. We separately report numeric results of two blur types: camera motion blur and defocus blur. We color code each row as \colorbox{best}{\textbf{best}} and \colorbox{second}{\textbf{second best}}. } 
\label{tab:ablation}
\end{table*}

\begin{figure*}
  \centering
  \rotatebox[origin=c]{90}{\makebox{\centering\scriptsize \hspace{10pt}Camera Motion Blur}}
  \begin{subfigure}{0.19\linewidth}
    \includegraphics[width=\linewidth]{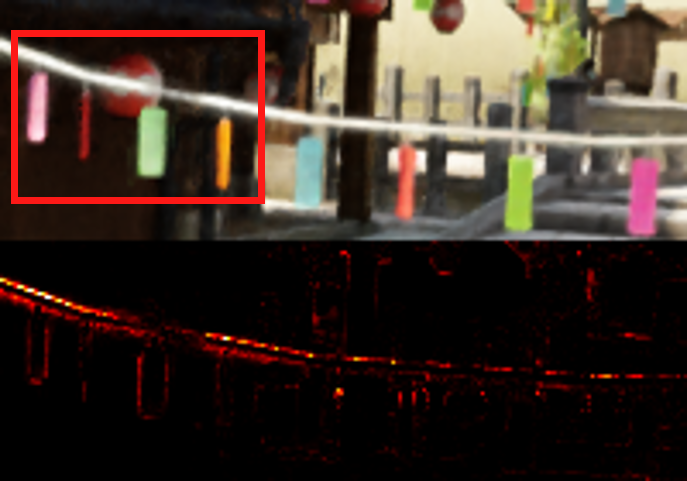}
    \label{subfig:ablation_wogamma}
  \end{subfigure}
  \hspace{-3pt}
  \begin{subfigure}{0.19\linewidth}
    \includegraphics[width=\linewidth]{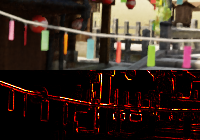}
    \label{subfig:ablation_woalign}
  \end{subfigure}
  \hspace{-3pt}
  \begin{subfigure}{0.19\linewidth}
    \includegraphics[width=\linewidth]{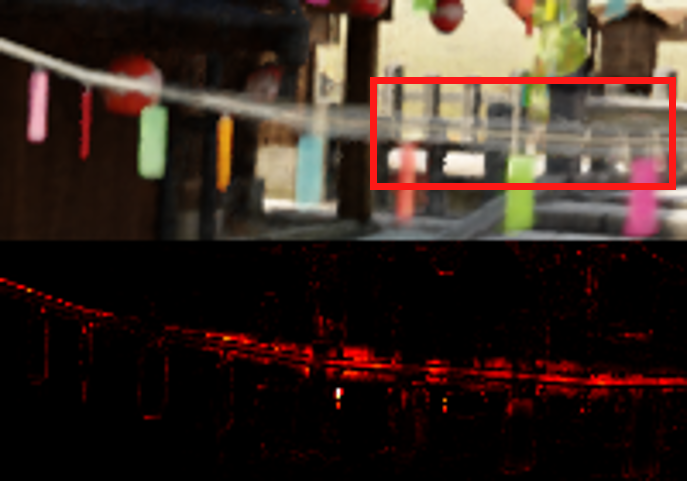}
    \label{subfig:ablation_wotrans}
  \end{subfigure}
  \hspace{-3pt}
  \begin{subfigure}{0.19\linewidth}
    \includegraphics[width=\linewidth]{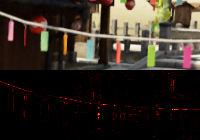}
    \label{subfig:ablation_full}
  \end{subfigure}
  \hspace{-3pt}
  \begin{subfigure}{0.19\linewidth}
    \includegraphics[width=\linewidth]{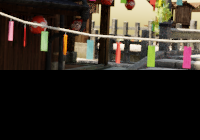}
    \label{subfig:ablation_gt}
  \end{subfigure}
  
  \vspace{\figtofig}
  \rotatebox[origin=c]{90}{\makebox{\centering \scriptsize \hspace{10pt}Defocus Blur}}
  \begin{subfigure}{0.19\linewidth}
    \includegraphics[width=\linewidth]{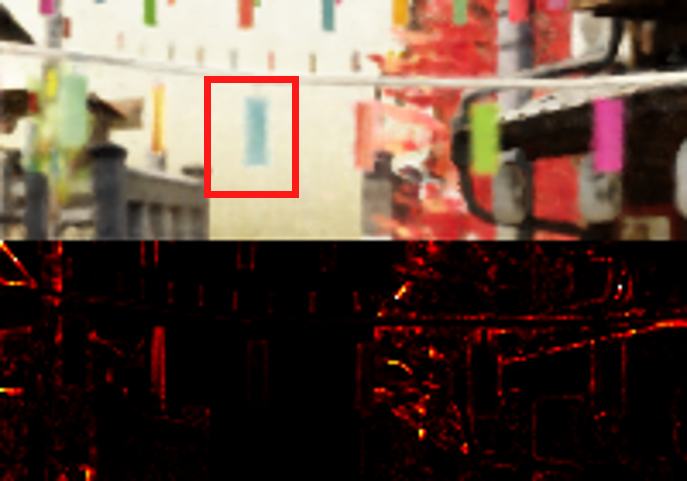}
    \vspace{\figtosubcap}
    \label{subfig:ablationdefocus_wogamma}
    \caption{w/o gamma}
  \end{subfigure}
  \hspace{-3pt}
  \begin{subfigure}{0.19\linewidth}
    \includegraphics[width=\linewidth]{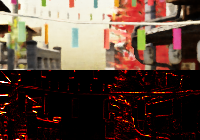}
    \vspace{\figtosubcap}
    \label{subfig:ablationdefocus_woalign}
    \caption{w/o align}
  \end{subfigure}
  \hspace{-3pt}
  \begin{subfigure}{0.19\linewidth}
    \includegraphics[width=\linewidth]{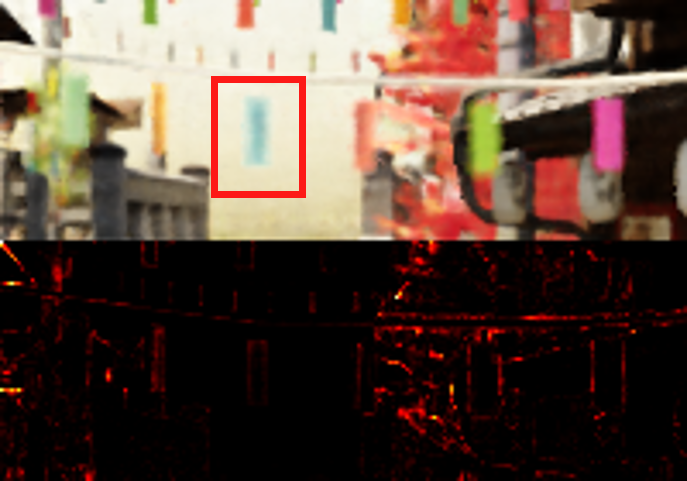}
    \vspace{\figtosubcap}
    \label{subfig:ablationdefocus_wotrans}
    \caption{w/o origin opt.}
  \end{subfigure}
  \hspace{-3pt}
  \begin{subfigure}{0.19\linewidth}
    \includegraphics[width=\linewidth]{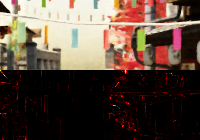}
    \vspace{\figtosubcap}
    \label{subfig:ablationdefocus_full}
    \caption{Ours}
  \end{subfigure}
  \hspace{-3pt}
  \begin{subfigure}{0.19\linewidth}
    \includegraphics[width=\linewidth]{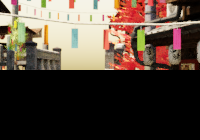}
    \vspace{\figtosubcap}
    \label{subfig:ablationdefocus_gt}
    \caption{Ground Truth}
  \end{subfigure}
  
  \vspace{\figtocapexp}
  \caption{Examples of ablations in the synthetic scenes. The corresponding error map is visualized in the bottom, where darker regions indicate smaller error. Our full model has the smallest error, especially at the edges. Note the artifacts highlighted in the red boxes.}
  \label{fig:ablation}
  \vspace{\captotext}
\end{figure*}

\tinysection{Datasets.} 
In the experiments we focus on two blur types: camera motion blur and defocus blur. For each type of blur we synthesize $5$ scenes using Blender \cite{misc_blender}. We manually place multi-view cameras to mimic real data capture. To render images with camera motion blur, we randomly perturb the camera pose, and then linearly interpolate poses between the original and perturbed poses for each view. We render images from interpolated poses and blend them in linear RGB space to generate the final blurry images. For defocus blur, we use the built-in functionality to render depth-of-field images. We fix the aperture and randomly choose a focus plane between the nearest and furthest depth. 

We also captured $20$ real world scenes with $10$ scenes for each blur type for a qualitative study. The camera used was a Canon EOS RP with manual exposure mode. We captured the camera motion blur images by manually shaking the camera during exposure, while the reference images are taken using a tripod. To capture defocus images, we choose a large aperture. We compute the camera poses of blurry and reference images in the real world scenes using the COLMAP \cite{misc_colmap1,misc_colmap2}. Although the estimated poses from COLMAP may be ambiguous due to the blur, we find our method is robust to inaccurate poses. One reason is that optimizing ray origins compensate for the registration errors.

\begin{figure}
  \centering
  \includegraphics[width=\linewidth]{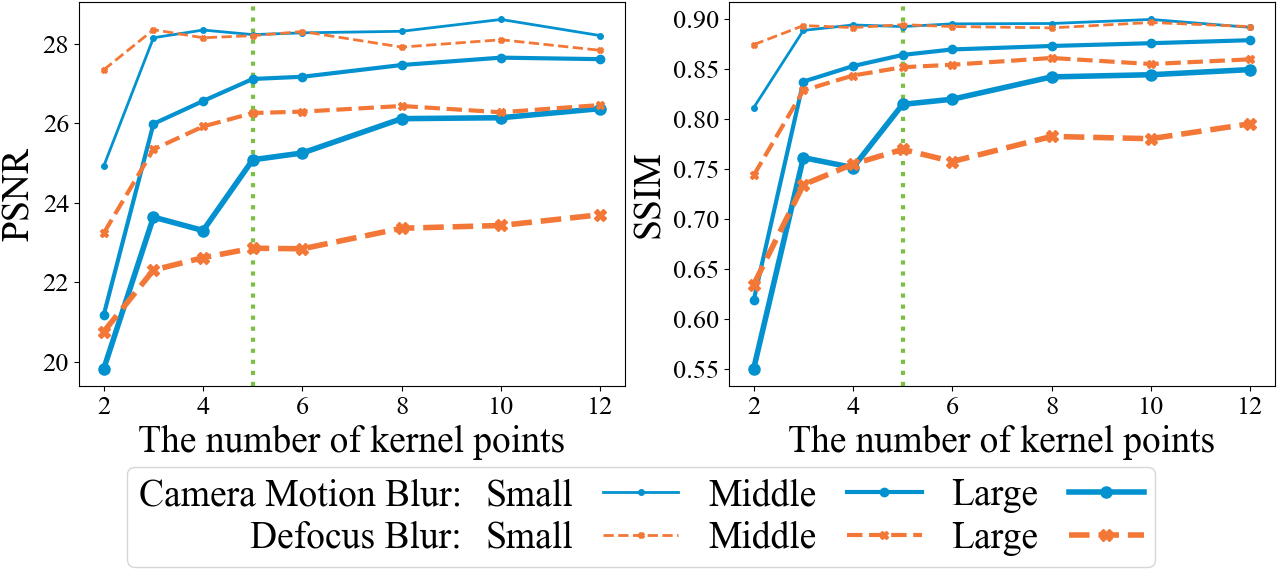}
  \vspace{-12pt}
  \caption{Comparison of our full model using a different number of kernel points. The vertical green lines indicate the $N = 5$, which we use in our other experiments. }
  \label{fig:ablation_ptnum}
  \vspace{-10pt}
\end{figure}

\subsection{Ablation Study}
\tinysection{Effectiveness of main components.} 
We first conduct ablations on several components in our framework: gamma correction (w/o gamma), ray origin optimization (w/o origin opt.), and the alignment loss (w/o align). We remove these components individually and train a separate NeRF in each of the synthetic camera motion blur and defocus blur datasets. We report the PSNR, SSIM, and LPIPS \cite{misc_lpips} metrics between the synthesized novel views and ground truth novel views. As shown in \cref{tab:ablation}, overall, the best result is achieved when the full model is used. We visualize the results and the error maps of two examples in the \cref{fig:ablation}. We note that other methods present larger errors, especially at the boundaries of objects. Without $\mathcal{L}_{align}$, the NeRF produces sharp but misaligned novel views.

\begin{table*}
\centering
\footnotesize
\setlength{\tabcolsep}{1pt}
\newcolumntype{Y}{>{\centering\arraybackslash}X}
\begin{tabularx}{0.999\linewidth}{l||YYY|YYY|YYY|YYY|YYY|YYY}
\toprule
& \multicolumn{3}{c}{\scshape Factory} 
 & \multicolumn{3}{c}{\scshape Cozyroom}  
 & \multicolumn{3}{c}{\scshape Pool} 
 & \multicolumn{3}{c}{\scshape Tanabata}
 & \multicolumn{3}{c}{\scshape Trolley} 
 & \multicolumn{3}{c}{\scshape Average} \\
  \textsl{Camera Motion} & \multicolumn{1}{c}{\scriptsize PSNR$\uparrow$}    & \multicolumn{1}{c}{\scriptsize SSIM$\uparrow$} & \multicolumn{1}{c}{\scriptsize LPIPS$\downarrow$} & \multicolumn{1}{c}{\scriptsize PSNR$\uparrow$} & \multicolumn{1}{c}{\scriptsize SSIM$\uparrow$} & \multicolumn{1}{c}{\scriptsize LPIPS$\downarrow$} & \multicolumn{1}{c}{\scriptsize PSNR$\uparrow$} & \multicolumn{1}{c}{\scriptsize SSIM$\uparrow$} & \multicolumn{1}{c}{\scriptsize LPIPS$\downarrow$} & \multicolumn{1}{c}{\scriptsize PSNR$\uparrow$}     & \multicolumn{1}{c}{\scriptsize SSIM$\uparrow$} & \multicolumn{1}{c}{\scriptsize LPIPS$\downarrow$} & \multicolumn{1}{c}{\scriptsize PSNR$\uparrow$}    & \multicolumn{1}{c}{\scriptsize SSIM$\uparrow$} & \multicolumn{1}{c}{\scriptsize LPIPS$\downarrow$} & \multicolumn{1}{c}{\scriptsize PSNR$\uparrow$}    & \multicolumn{1}{c}{\scriptsize SSIM$\uparrow$} & \multicolumn{1}{c}{\scriptsize LPIPS$\downarrow$}  \\
  \midrule
\textit{naive \scriptsize NeRF}             & 19.32   & .4563    & .5304     & 25.66    & .7941    & .2288     & \cellcolor{third}30.45    & \cellcolor{third}.8354    & \cellcolor{third}.1932     & 22.22    & .6807    & .3653     & 21.25   & .6370    & .3633     & 23.78   & .6807    & .3362      \\
\scriptsize \textit{MPR + NeRF}               & \cellcolor{second}21.70       & \cellcolor{second}.6153    & \cellcolor{second}.3094     & \cellcolor{second}27.88    & \cellcolor{second}.8502    & \cellcolor{second}.1153     & \cellcolor{second}30.64    & \cellcolor{second}.8385    & \cellcolor{second}.1641     & \cellcolor{third}22.71        & \cellcolor{third}.7199    & \cellcolor{second}.2509     & \cellcolor{third}22.64       & \cellcolor{third}.7141    & \cellcolor{second}.2344     & \cellcolor{second}25.11       & \cellcolor{second}.7476    & \cellcolor{second}.2148      \\
\scriptsize \textit{PVD + NeRF}               & \cellcolor{third}20.33       & \cellcolor{third}.5386    & \cellcolor{third}.3667     & \cellcolor{third}27.74    & \cellcolor{third}.8296    & \cellcolor{third}.1451     & 27.56    & .7626    & .2148     & \cellcolor{second}23.44        & \cellcolor{second}.7293    & \cellcolor{third}.2542     & \cellcolor{second}23.81       & \cellcolor{second}.7351    & \cellcolor{third}.2567     & \cellcolor{third}24.58       & \cellcolor{third}.7190    & \cellcolor{third}.2475      \\
Ours              & \cellcolor{best}25.60       & \cellcolor{best}.7750    & \cellcolor{best}.2687     & \cellcolor{best}32.08    & \cellcolor{best}.9261    & \cellcolor{best}.0477     & \cellcolor{best}31.61    & \cellcolor{best}.8682    & \cellcolor{best}.1246     & \cellcolor{best}27.11        & \cellcolor{best}.8640    & \cellcolor{best}.1228     & \cellcolor{best}27.45       & \cellcolor{best}.8632    & \cellcolor{best}.1363     & \cellcolor{best}28.77       & \cellcolor{best}.8593    & \cellcolor{best}.1400      \\
\midrule
\midrule
\textsl{Defocus}  & \multicolumn{1}{c}{\scriptsize PSNR$\uparrow$}    & \multicolumn{1}{c}{\scriptsize SSIM$\uparrow$} & \multicolumn{1}{c}{\scriptsize LPIPS$\downarrow$} & \multicolumn{1}{c}{\scriptsize PSNR$\uparrow$} & \multicolumn{1}{c}{\scriptsize SSIM$\uparrow$} & \multicolumn{1}{c}{\scriptsize LPIPS$\downarrow$} & \multicolumn{1}{c}{\scriptsize PSNR$\uparrow$} & \multicolumn{1}{c}{\scriptsize SSIM$\uparrow$} & \multicolumn{1}{c}{\scriptsize LPIPS$\downarrow$} & \multicolumn{1}{c}{\scriptsize PSNR$\uparrow$} & \multicolumn{1}{c}{\scriptsize SSIM$\uparrow$} & \multicolumn{1}{c}{\scriptsize LPIPS$\downarrow$} & \multicolumn{1}{c}{\scriptsize PSNR$\uparrow$}    & \multicolumn{1}{c}{\scriptsize SSIM$\uparrow$} & \multicolumn{1}{c}{\scriptsize LPIPS$\downarrow$} & \multicolumn{1}{c}{\scriptsize PSNR$\uparrow$}    & \multicolumn{1}{c}{\scriptsize SSIM$\uparrow$} & \multicolumn{1}{c}{\scriptsize LPIPS$\downarrow$}  \\
\midrule
\textit{naive \scriptsize NeRF}  & \cellcolor{third}25.36       & \cellcolor{third}.7847    & \cellcolor{third}.2351     & \cellcolor{second}30.03    & \cellcolor{second}.8926    & \cellcolor{third}.0885    & \cellcolor{second}27.77    & \cellcolor{second}.7266    & \cellcolor{third}.3340     & \cellcolor{third}23.80        & \cellcolor{third}.7811    & \cellcolor{third}.2142     & \cellcolor{third}22.67       & \cellcolor{third}.7103    & \cellcolor{third}.2799     & \cellcolor{second}25.93       & \cellcolor{third}.7791    & \cellcolor{third}.2303      \\
\scriptsize \textit{KPAC + NeRF}  & \cellcolor{second}26.40       & \cellcolor{second}.8194    & \cellcolor{second}.1624     & \cellcolor{third}28.15    & \cellcolor{third}.8592    & \cellcolor{second}.0815     & \cellcolor{third}26.69    & \cellcolor{third}.6589    & \cellcolor{second}.2631     & \cellcolor{second}24.81        & \cellcolor{second}.8147    & \cellcolor{second}.1639     & \cellcolor{second}23.42       & \cellcolor{second}.7495    & \cellcolor{second}.2155     & \cellcolor{third}25.89       & \cellcolor{second}.7803    & \cellcolor{second}.1773      \\
Ours & \cellcolor{best}28.03       & \cellcolor{best}.8628    & \cellcolor{best}.1127     & \cellcolor{best}31.85    & \cellcolor{best}.9175    & \cellcolor{best}.0481     & \cellcolor{best}30.52    & \cellcolor{best}.8246    & \cellcolor{best}.1901     & \cellcolor{best}26.25        & \cellcolor{best}.8517    & \cellcolor{best}.0995     & \cellcolor{best}25.18       & \cellcolor{best}.8067    & \cellcolor{best}.1436     & \cellcolor{best}28.37       & \cellcolor{best}.8527    & \cellcolor{best}.1188     \\
\bottomrule
\end{tabularx}
\vspace{-0.3cm}
\caption{Quantitative comparison on synthetic scenes of two blur types. We color code each row as \colorbox{best}{\textbf{best}} and \colorbox{second}{\textbf{second best}}}
\vspace{-0.2cm}
\label{tab:comparsion}
\end{table*}

\begin{figure*}
\newlength{\blurw}
\newlength{\defocusw}

\setlength{\blurw}{0.19\linewidth}
\setlength{\defocusw}{0.24\linewidth}
  \centering
  
  \hfill
  \begin{subfigure}{\blurw}
    \includegraphics[width=\linewidth]{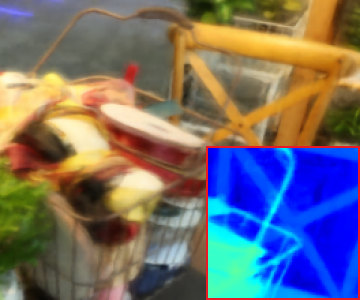}
    \label{subfig:comparison_blurbasket_naive}
  \end{subfigure}
  \hfill
  \begin{subfigure}{\blurw}
    \includegraphics[width=\linewidth]{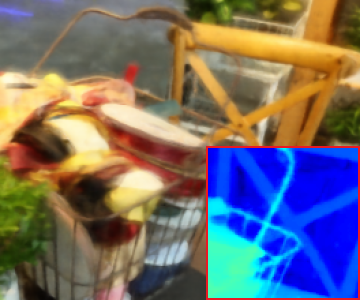}
    \label{subfig:comparison_blurbasket_mpr}
  \end{subfigure}
  \hfill
  \begin{subfigure}{\blurw}
    \includegraphics[width=\linewidth]{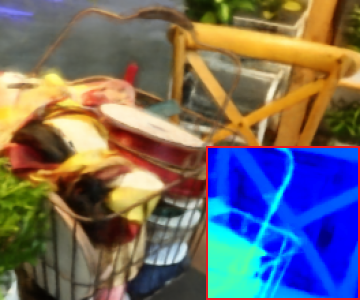}
    \label{subfig:comparison_blurbasket_pvd}
  \end{subfigure}
  \hfill
  \begin{subfigure}{\blurw}
    \includegraphics[width=\linewidth]{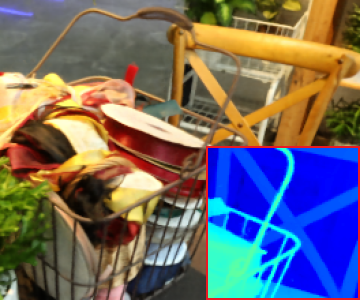}
    \label{subfig:comparison_blurbasket_full}
  \end{subfigure}
  \hfill
  \begin{subfigure}{\blurw}
    \includegraphics[width=\linewidth]{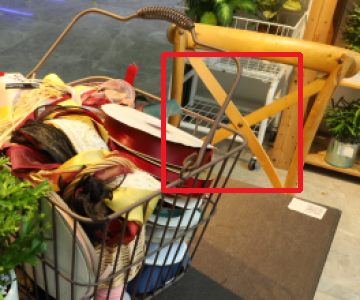}
    \label{subfig:comparison_blurbasket_gt}
  \end{subfigure}
  \hfill
  
  \vspace{\figtofig}
  \hfill
  \begin{subfigure}{\blurw}
    \includegraphics[width=\linewidth]{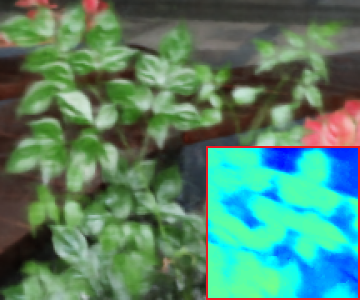}
    \label{subfig:comparison_blurparterre_naive}
  \end{subfigure}
  \hfill
  \begin{subfigure}{\blurw}
    \includegraphics[width=\linewidth]{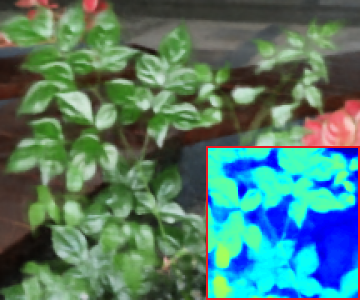}
    \label{subfig:comparison_blurparterre_mpr}
  \end{subfigure}
  \hfill
  \begin{subfigure}{\blurw}
    \includegraphics[width=\linewidth]{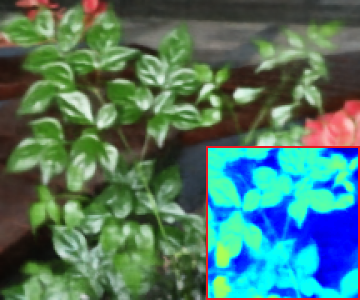}
    \label{subfig:comparison_blurparterre_pvd}
  \end{subfigure}
  \hfill
  \begin{subfigure}{\blurw}
    \includegraphics[width=\linewidth]{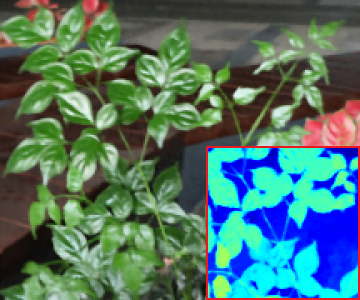}
    \label{subfig:comparison_blurparterre_full}
  \end{subfigure}
  \hfill
  \begin{subfigure}{\blurw}
    \includegraphics[width=\linewidth]{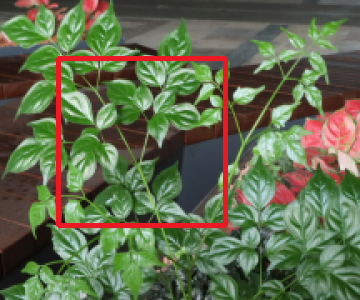}
    \label{subfig:comparison_blurparterre_gt}
  \end{subfigure}
  \hfill
  
  \vspace{\figtofig}
  \hfill
  \begin{subfigure}{\blurw}
    \includegraphics[width=\linewidth]{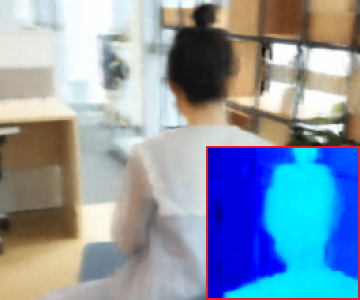}
    \label{subfig:comparison_blurgirl_naive}
    \vspace{\figtosubcap}
    \caption*{\textit{naive NeRF}}
  \end{subfigure}
  \hfill
  \begin{subfigure}{\blurw}
    \includegraphics[width=\linewidth]{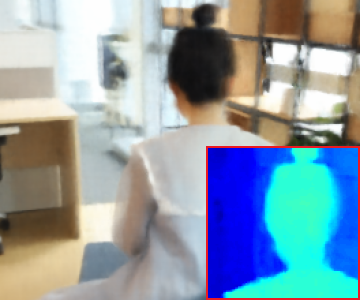}
    \label{subfig:comparison_blurgirl_mpr}
    \vspace{\figtosubcap}
    \caption*{\textit{MPR + NeRF}}
  \end{subfigure}
  \hfill
  \begin{subfigure}{\blurw}
    \includegraphics[width=\linewidth]{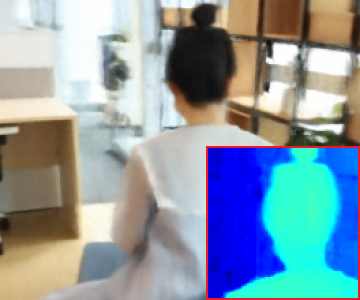}
    \label{subfig:comparison_blurgirl_pvd}
    \vspace{\figtosubcap}
    \caption*{\textit{PVD + NeRF}}
  \end{subfigure}
  \hfill
  \begin{subfigure}{\blurw}
    \includegraphics[width=\linewidth]{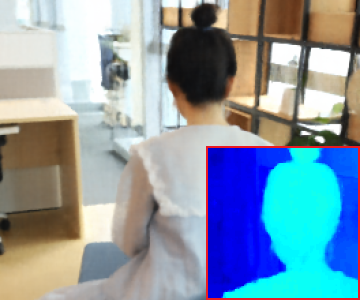}
    \label{subfig:comparison_blurgirl_full}
    \vspace{\figtosubcap}
    \caption*{Ours}
  \end{subfigure}
  \hfill
  \begin{subfigure}{\blurw}
    \includegraphics[width=\linewidth]{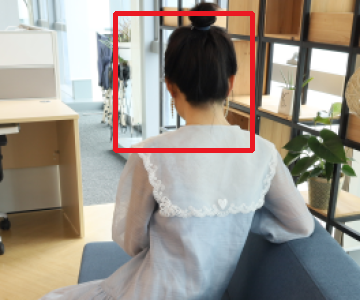}
    \label{subfig:comparison_blurgirl_gt}
    \vspace{\figtosubcap}
    \caption*{Reference}
  \end{subfigure}
  \hfill

  \vspace{\figtocapcamp}
  \caption{Qualitative comparison on real world camera motion blur. The last column is captured for reference only and may be misaligned with the ground truth.}
  \label{fig:comparison}
  \vspace{-0.5cm}
\end{figure*}

\tinysection{Number of kernel points.}
One important hyper-parameter in our method is the number of sparse locations $N = |\mathcal{N}(\mathbf{p})|$. We experimented with different values of $N$ with our full model in various blur situations, considering both blur types and three degrees of blur. We plot the PSNR and SSIM curves in \cref{fig:ablation_ptnum}. We note that in all cases the quality of the results improves as $N$ increases. However, the improvement is less substantial beyond $N = 5$. In case the input is extremely blurry, further increasing the number of kernel points can potentially help. However, increasing $N$ comes at a larger computation and memory cost during training. Therefore, we use $N = 5$ for all other experiments, providing a good balance between rendering quality and efficiency.


\begin{figure*}
\setlength{\blurw}{0.24\linewidth}
\setlength{\figtofig}{-11pt}
  \centering
  
  \hfill
  \begin{subfigure}{\blurw}
    \includegraphics[width=\linewidth]{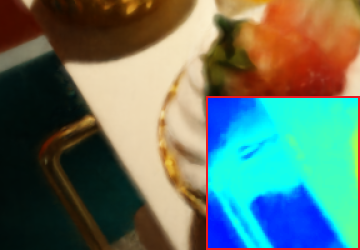}
    \label{subfig:comparison_defocuscupcake_naive}
  \end{subfigure}
  \hfill
  \begin{subfigure}{\blurw}
    \includegraphics[width=\linewidth]{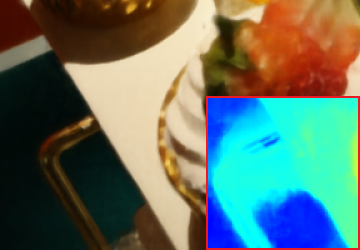}
    \label{subfig:comparison_defocuscupcake_kpac}
  \end{subfigure}
  \hfill
  \begin{subfigure}{\blurw}
    \includegraphics[width=\linewidth]{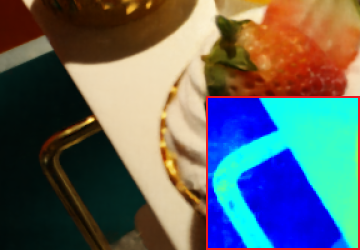}
    \label{subfig:comparison_defocuscupcake_full}
  \end{subfigure}
  \hfill
  \begin{subfigure}{\blurw}
    \includegraphics[width=\linewidth]{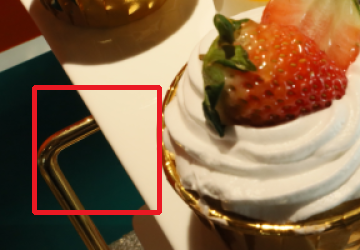}
    \label{subfig:comparison_defocuscupcake_gt}
  \end{subfigure}
  \hfill
  
  \vspace{\figtofig}
  \hfill
  \begin{subfigure}{\blurw}
    \includegraphics[width=\linewidth]{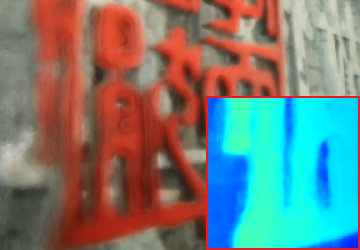}
    \label{subfig:comparison_defocusseal_naive}
  \end{subfigure}
  \hfill
  \begin{subfigure}{\blurw}
    \includegraphics[width=\linewidth]{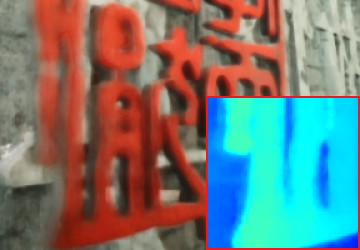}
    \label{subfig:comparison_defocusseal_kpac}
  \end{subfigure}
  \hfill
  \begin{subfigure}{\blurw}
    \includegraphics[width=\linewidth]{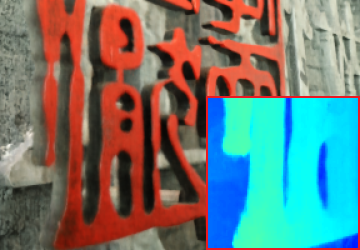}
    \label{subfig:comparison_defocusseal_full}
  \end{subfigure}
  \hfill
  \begin{subfigure}{\blurw}
    \includegraphics[width=\linewidth]{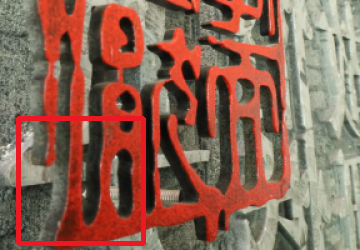}
    \label{subfig:comparison_defocusseal_gt}
  \end{subfigure}
  \hfill
  
  \vspace{\figtofig}
  \hfill
  \begin{subfigure}{\blurw}
    \includegraphics[width=\linewidth]{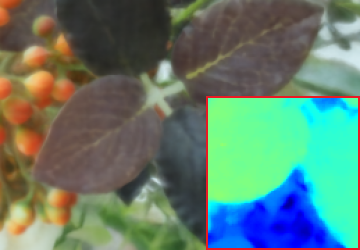}
    \label{subfig:comparison_defocuscoral_naive}
    \vspace{\figtosubcap}
    \caption*{\textit{naive NeRF}}
  \end{subfigure}
  \hfill
  \begin{subfigure}{\blurw}
    \includegraphics[width=\linewidth]{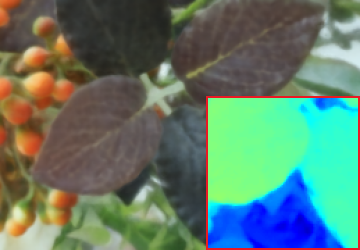}
    \label{subfig:comparison_defocuscoral_kpac}
    \vspace{\figtosubcap}
    \caption*{\textit{KPAC + NeRF}}
  \end{subfigure}
  \hfill
  \begin{subfigure}{\blurw}
    \includegraphics[width=\linewidth]{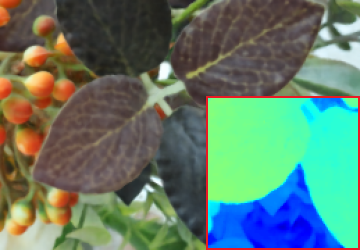}
    \label{subfig:comparison_defocuscoral_full}
    \vspace{\figtosubcap}
    \caption*{Ours}
  \end{subfigure}
  \hfill
  \begin{subfigure}{\blurw}
    \includegraphics[width=\linewidth]{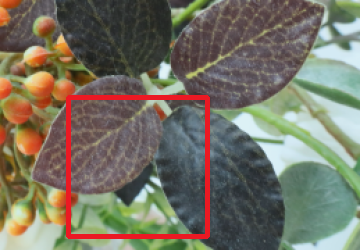}
    \label{subfig:comparison_defocuscoral_gt}
    \vspace{\figtosubcap}
    \caption*{Reference}
  \end{subfigure}
  \hfill
  
  \vspace{\figtocapcamp}
  \caption{Qualitative comparison on real world defocus blur. The last column is captured for reference only and may be misaligned or have different exposures than ground truth.}
  \vspace{-0.5cm}
  \label{fig:comparison_defoucs}
\end{figure*}

\subsection{Comparisons}
Since there are no existing works that try to reconstruct the NeRF from blurry input for novel view synthesis, we carefully select several possible baselines to compare with. The most straightforward one would be to directly train the NeRF using the blurry input (\textit{naive NeRF}). Additionally, we also compare to the image-space baselines that first restore the input using the existing image or video based deblurring techniques and then train the NeRF with the deblurred images. For camera motion blur, we compare with the current state-of-the-art methods for deblurring on single image \cite{Blur_CNN_MPR} (\textit{MPR + NeRF}) and video \cite{Blur_video_PVD} (\textit{PVD + NeRF}). For defocus blur, we compare to the KPAC \cite{Defocus_KPAC} (\textit{KPAC + NeRF}). We show quantitative results on synthetic scenes in \cref{tab:comparsion}. For real scenes, due to the nature of capturing, the ground truth images are not available due to the misalignment for camera motion blur or exposure variations for defocus blur. We can see that while the image-space deblurring baseline improves compared to the \textit{naive NeRF} baseline, our full pipeline outperforms these baselines to a large extent in the synthetic scenes among two blur types. The video based deblurring method \textit{PVD + NeRF} sometimes performs worse than the single-image based method \textit{MPR + NeRF}. A possible reason is that \textit{PVD + NeRF} aggregates features from neighbor frames based on optical flow, which is really challenging for blurry input with large baselines. \cref{fig:comparison} and~\ref{fig:comparison_defoucs} showcase the qualitative comparison results on camera motion blur and defocus blur, respectively, in real scenes. Our method produces novel views with sharp edges and rich details, being the closest to ground truth. Previous methods demonstrate artifacts near the object boundary and blurry textures. The depth maps we predict are also sharper compared to other methods. Moreover, our method could produce more view consistent results than other baselines. And we provide additional results in conjunction with video output in the \textit{supplementary material}.

\section{Discussion and Conclusion}
\subsection{Why our framework works}
Blindly recovering a sharp NeRF and the blur kernel simultaneously with only blurry images is an ill-posed problem, as the NeRF can also reconstruct a blurry scene that may ``explain'' the blurry images. Then how does our framework ensures that we get a sharp NeRF?
As illustrated in the NeRF++ \cite{NeRF_misc_plusplus}, the NeRF encodes priors for view consistent reconstruction. When the blurry input is view inconsistent, our framework compensates for the inconsistency using the DSK module, leading to the decomposition of the consistent sharp scene and the inconsistent blur pattern.
Note that we define the blurry input as view consistent if they are equivalent to sharp observations of one blurry 3D scene.
Blur in the real world is usually inconsistent. Each shot has a different blur pattern due to the randomness of camera movement or variability of focus distance. This can be further validated by the fact that when the \textit{naive NeRF} reconstructs the real scene dataset, the results flicker severely when the viewpoint changes. Our framework addresses this issue, which is common in real world data.

\subsection{Limitations}
Our method can fail when the blur is view consistent, e.g., the camera coincidentally shaking in roughly the same direction across all views, or the camera having a fixed focal point (i.e., focuses on a single target). 
Deblurring a consistent blur can potentially be solved by introducing image priors, which we treat as future work.
Our method may also fail when encounter input images that are severely blurred because the COLMAP may fail to reconstruct the camera poses. But in the experiments we find that this is only an issue in very blurry cases. For further discussion about such limitations, please refer to the \textit{supplementary material}.


\subsection{Conclusion}
In this paper, we propose a simple but effective framework for training a sharp NeRF under blurry input. Experiments on both synthetic and real world scenes verify the effectiveness of our framework and demonstrate the significant improvement in quality over \textit{naive NeRF} and image-space deblurring approaches. We hope that this work will further motivate research into NeRF-based approaches for deblurring applications. 

\tinysection{Acknowledgements.} Authors at HKUST and CityU were partly supported by the Hong Kong Research Grants Council (RGC), including the RGC Early Career Scheme under Grant 9048148 (CityU 21209119).

{\small
\bibliographystyle{ieee_fullname}
\bibliography{egbib}
}

\end{document}